\documentclass[lettersize,journal]{IEEEtran}
\usepackage{amsmath,amsfonts}
\usepackage{algorithm}
\usepackage{array}
\usepackage[caption=false,font=normalsize,labelfont=sf,textfont=sf]{subfig}
\usepackage{textcomp}
\usepackage{stfloats}
\usepackage{url}
\usepackage{verbatim}
\usepackage{graphicx}
\usepackage{cite}

\usepackage{booktabs}
\usepackage{times}
\usepackage{amsmath,amssymb}
\usepackage{empheq}
\usepackage{graphicx}
\usepackage{wrapfig}
\usepackage{sidecap}
\usepackage{soul}
\usepackage{enumitem}
\usepackage{setspace}
\usepackage{parskip}
\usepackage{xcolor}
\usepackage{tikz}
\usepackage{float}

\usepackage{graphicx}
\usepackage{multirow}
\usepackage{array}
\usepackage{colortbl}
\usepackage[table,xcdraw]{xcolor}
\usepackage[dvipsnames]{xcolor}
\usepackage{makecell}
\usepackage{algpseudocode}

\usepackage{booktabs}
\usepackage{pifont}      
\usepackage{amssymb}     
\usepackage{colortbl}    
\usepackage{xcolor}      
\usepackage{multirow}

\definecolor{Gray}{gray}{0.8}
\definecolor{blue}{rgb}{0.21,0.49,0.74}
\usepackage[pagebackref,breaklinks,colorlinks,allcolors=blue]{hyperref}
\usepackage[capitalize]{cleveref}
\crefname{section}{Sec.}{Secs.}
\Crefname{section}{Section}{Sections}
\Crefname{table}{Table}{Tables}
\crefname{table}{Tab.}{Tabs.}

\begin{document}
     
\title{Personalized Generative Low-light Image \\Denoising and Enhancement}



\author{%
Xijun Wang\IEEEauthorrefmark{1},~\IEEEmembership{Member,~IEEE,} 
Prateek Chennuri\IEEEauthorrefmark{1},~\IEEEmembership{Student Member,~IEEE,} 
Dilshan Godaliyadda\IEEEauthorrefmark{2}, ~\IEEEmembership{Member,~IEEE,} 
Yu Yuan\IEEEauthorrefmark{1},~\IEEEmembership{Student Member,~IEEE,} 
Bole Ma\IEEEauthorrefmark{1},~\IEEEmembership{Student Member,~IEEE,} 
Xingguang Zhang\IEEEauthorrefmark{1},~\IEEEmembership{Student Member,~IEEE,} 
Hamid R. Sheikh\IEEEauthorrefmark{2}, ~\IEEEmembership{Fellow,~IEEE,} 
Stanley H. Chan\IEEEauthorrefmark{1},~\IEEEmembership{Senior Member,~IEEE}\\
\IEEEauthorrefmark{1}School of Electrical and Computer Engineering, Purdue University, West Lafayette, IN, USA\\
\IEEEauthorrefmark{2}Samsung Research America, Plano, TX, USA
\thanks{This work is supported, in part, by a research contract from Samsung Research America, NSF IIS-2133032, and NSF ECCS-2030570.}%
}

\markboth{Journal of \LaTeX\ Class Files,~Vol.~14, No.~8, August~2021}%
{Shell \MakeLowercase{\textit{et al.}}: Personalized Generative Low-light Image Denoising and Enhancement}


\maketitle

\begin{abstract}
Modern cameras' performance in low-light conditions remains suboptimal due to fundamental limitations in photon shot noise and sensor read noise. Generative image restoration methods have shown promising results compared to traditional approaches, but they suffer from hallucinatory content generation when the signal-to-noise ratio (SNR) is low. Leveraging the availability of personalized photo galleries of the users, we introduce Diffusion-based Personalized Generative Denoising (DiffPGD), a new approach that builds a customized diffusion model for individual users. Our key innovation lies in the development of an identity-consistent physical buffer that extracts the physical attributes of the person from the gallery. This ID-consistent physical buffer serves as a robust prior that can be seamlessly integrated into the diffusion model to restore degraded images without the need for fine-tuning. Over a wide range of low-light testing scenarios, we show that DiffPGD achieves superior image denoising and enhancement performance compared to existing diffusion-based denoising approaches. Our project page can be found at \href{https://genai-restore.github.io/DiffPGD/}{\textcolor{purple}{\textbf{https://genai-restore.github.io/DiffPGD/}}}. 
\end{abstract}


\section{Introduction}
\label{sec:intro}

With the astonishing development of smartphones, it is safe to say that smartphone cameras today have surpassed digital single-lens reflex (DSLR) cameras in both popularity and functionality. However, the small form factor of smartphone cameras puts a tight constraint on the aperture size, hence limiting the amount of light a CMOS pixel can detect. In a low-light imaging environment, this creates a fundamental limitation to how short the exposure can be and how much signal-to-noise ratio (SNR) the sensor can support. While some of the noise seen at low light can be mitigated using better CMOS technology (e.g., correlated double sampling to reduce the read noise \cite{Nakamura2021sensor} and deep-well pump gate to reduce the stray capacitance \cite{ma2015pump, Ma2022qis}), the Poisson statistics due to random photon arrivals is a problem created by mother nature that cannot be solved by even the ideal sensors. Therefore, image denoising and enhancement by exploiting the internal structures of images become a necessary task for all smartphone cameras.

From a signal-processing perspective, degradation caused by low light can be roughly approximated via
\begin{equation}\label{eq:poisson_gaussian}
    \mathbf{y} = \text{ADC}\big( \text{Poisson}(\alpha \mathbf{x} + \eta)\big) + \text{Gauss}(0, \sigma_{\text{read}}^2),
\end{equation}
where $\mathbf{x}$ is the unknown clean image, $\alpha$ is the quantum efficiency, $\eta$ is the dark current, $\mathrm{ADC}(\cdot)$ models the analog-to-digital conversion that quantizes the sensor’s analog voltage into digital pixel values, and $\text{Gauss}(0, \sigma_{\text{read}}^2)$ is the read noise. The inverse problem associated with low-light denoising is to recover $\mathbf{x}$ from $\mathbf{y}$. Over the past five decades, the main driving force is to exploit image structures, from Wiener filter \cite{Lim1989TwoDimensionalSA}, total variation \cite{Rudin1994TV, Chambolle2004tv}, wavelets \cite{Coifman1995} to non-local means \cite{Buades2005nl}, BM3D \cite{Dabov2007BM3D}, and recently convolution neural networks \cite{Jin2017cnn, Guo2019real, wei2020physics, chi2020dynamic, Elgendy2021LowLightDA}, transformers \cite{Liang2021swinir, Lu2022low, zhang2024xformer}, and diffusion models \cite{kawar2022denoising, zhu2023DiffPIR, Wang2025GenerativePB, Yuan2025iHDRIH}. A common theme across existing methods is to find and utilize a \emph{generic} prior $p(\mathbf{x})$, i.e., a prior learned from a large collection of example images. While they perform well in moderately difficult problems, the generality of these methods cannot be extended to heavily corrupted images. In the context of human face recovery, the restored results often lack real identity and exhibit artifacts. This is attributed to the ill-posed nature of the problem and the lack of appropriate constraints.

\begin{figure}[h]
\centering
\includegraphics[width=1.0\linewidth]{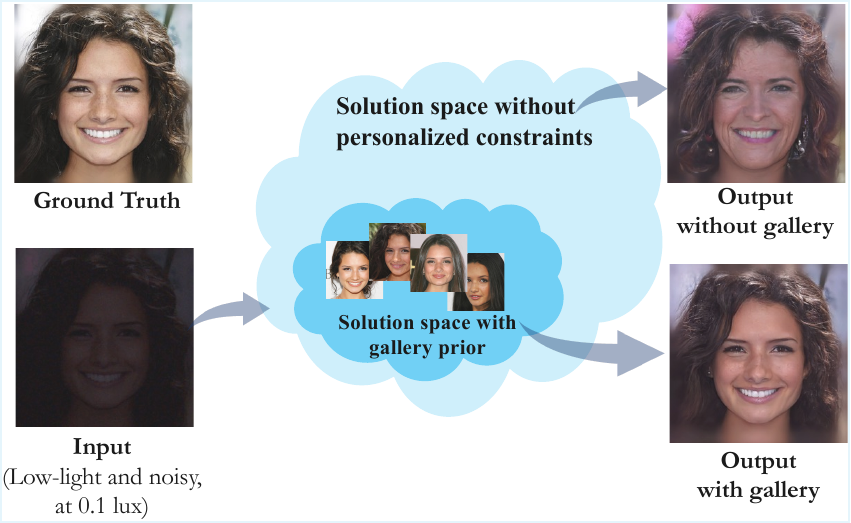}
\caption{The restoration of inputs degraded by noise and low-light conditions is highly ill-posed. By incorporating additional high-quality gallery photos of the same identity, we significantly reduce the solution space, thereby achieving improved identity consistency in the restored images.}
\label{fig:idea}
\end{figure}

\textbf{Why Gallery Photos?} Smartphone cameras today typically store hundreds if not thousands of a user's photos, captured at different times, in different places, and under different lighting conditions. While these images have many variations, they are all about the same person(s). Therefore, if the imaging goal is to take a photo of this user, the gallery on the phone would be the best source to build a prior $p(\mathbf{x})$. This concept is illustrated in Fig.~\ref{fig:idea}. In this paper, we are interested in diffusion-based image restoration. The role of gallery is to provide a strong constraint to the search problem. This allows us to search for better quality images and preserve the person's identity.

There are, however, two technical questions: 
\begin{itemize}
\item What class of images we should focus on: In this paper, we focus on \emph{human faces} because of their relevance to phone users. We assume that the gallery photos have been selected and are reasonably informative. Product-level engineering such as pre-processing and selection of the gallery photos is important but beyond the scope of this paper. 
\item How to incorporate the class-specific $p(\mathbf{x})$ into the restoration model: This is a bigger (and harder) question. Given the gallery photos, how do we efficiently extract the prior information and improve the restoration? We do not want to train a restoration model from scratch because the gallery can be small. We also want to avoid fine-tuning during inference. Moreover, the person's identity needs to be preserved. 
\end{itemize}

\textbf{Physical Buffers to the Rescue}. Given the gallery photos, what kind of prior information would be useful for restoration? Advancements in computer vision have made it possible to extract facial \emph{physical buffers}—including albedo and normal maps—from a person's gallery of images\cite{roth2016adaptive, Zhang2021lap,suwajanakorn2015makes}. These physical buffers capture crucial identity-defining properties such as surface geometry and skin color, effectively encoding an individual's unique identity. At the same time, it also eliminates the external influence of environmental lighting, pose, and other identity-independent variables. By incorporating this robust prior that disentangles the external influences, our approach can effectively extract identity information from the gallery.

\textbf{Contributions.} Our contributions are as follows:
\begin{enumerate}
    \item We introduce a personalized low-light denoising framework based on diffusion models. The proposed framework extracts person-specific priors from gallery images and integrates them into the diffusion process to effectively restore low-light facial images. Our solution offers three advantages: (i) We outperform existing methods in low-light conditions. (ii) We preserve person identity better than existing methods. (iii) We do not need to fine-tune for individuals.
    \item We propose to leverage physical buffer as an explicit prior for identity-preserving restoration. By learning the physical prior from clean data rather than noisy observations, the model captures intrinsic and noise-independent identity features. This physically grounded prior demonstrates strong generalization capability—models trained on synthetic data can effectively transfer to real-world scenarios.
\end{enumerate}

\section{Related Works}
\label{sec:related_works}

\textbf{Low-light Image Enhancement and Denoising.} Low-light images often suffer from significant noise due to low signal-to-noise ratios (SNR), in addition to reduced brightness and contrast. Traditional low-light denoising techniques often focus on exploiting image structures, ranging from early methods like Wiener filtering \cite{Lim1989TwoDimensionalSA}, total variation \cite{Rudin1994TV, Chambolle2004tv}, and wavelets \cite{Coifman1995} to non-local means \cite{Buades2005nl} and BM3D \cite{Dabov2007BM3D}. More recently, deep learning approaches, including convolutional neural networks \cite{Jin2017cnn, Guo2019real, wei2020physics, zheng2021adaptive, Fang2025SRENetSL, chi2020dynamic, Elgendy2021LowLightDA}, transformers \cite{Liang2021swinir, Lu2022low, zhang2024xformer, Yang2025LearningTS}, and diffusion-based models \cite{kawar2022denoising, zhu2023DiffPIR, Nguyen2024diffusion, CheukYiuChan2024AnlightenDiffAD, Wu2025MutuallyRL, Prateek2025quantadiff}, have emerged as promising solutions for low-light denoising. Low-light image enhancement (LLIE) aims to improve the brightness and contrast of images captured in poorly lit conditions. Traditional approaches \cite{Fu2016Weighted, Li2018Structure, Lee2013Contrast} often leverage heuristic or physical models to enhance image quality, while learning-based methods \cite{Land1971Retinex, Chen2018Retinex, Jiang2021EnlightenGAN, Chen2018See, Guo2020Zero, Lu2021Tbefn, wang2023exposurediffusion, xu2022snr, jiang2023low} have demonstrated greater effectiveness when supported by large-scale datasets. Some works also adopt training-free (zero-shot) learning for LLIE \cite{Lv2024FourierPD, Huang2025ZeroShotLI, Xie2024ResidualQL}. 
Recent advancements \cite{Shi2024ZERO, Lu2022low, Nguyen2024diffusion, xu2020learning} in this field have increasingly focused on jointly addressing denoising and enhancement tasks, aiming to improve image quality while effectively suppressing noise to maintain natural appearance. However, to the best of our knowledge, no specific approach currently exists for joint enhancement and denoising of facial images in low-light conditions.

\textbf{Deep Face Restoration.} CNN/Transformer-based face restoration methods \cite{wang2022RestoreFormer, Fan2022faceformer, Jiang2022dual}, such as RestoreFormer \cite{wang2022RestoreFormer}, typically rely on paired low- and high-quality image datasets to learn resolution enhancement and noise reduction. CodeFormer \cite{zhou2022towards} formulates face restoration as a code prediction task in a discrete latent space, leveraging a Transformer to recover identity-consistent and perceptually realistic results. While effective, these methods often struggle under severe degradation and lack the capability to synthesize realistic details in complex scenarios. 

In recent years, generative priors have been increasingly utilized in face restoration to better address real-world degradations. GAN-based methods, such as GFP-GAN \cite{wang2021towards}, leverage StyleGAN2 \cite{tov2021stylegan2} to enhance facial details, while diffusion-based models \cite{wang2023dr2, Chen2024face, ding2024restoration, Tao2024face, Varanka2024gallery} have further expanded restoration capabilities. For instance, BFRDiffusion \cite{Chen2024face} uses Stable Diffusion to enhance low-quality images by adding high-fidelity details, and Ding \textit{et al.} \cite{ding2024restoration} employ a pretrained diffusion model that denoises degraded inputs while preserving identity by fine-tuning on a person's gallery photos. MGBFR \cite{Tao2024face} combines text prompts and reference images with a dual-control adapter to retain accurate facial attributes and identity. Although these generative methods are effective, they often rely on multi-stage training or fine-tuning and often struggle recovering identity-consistent face under severe degradation,  which can limit the efficient use in some real-world applications.

\textbf{Reference Prior for Face Restoration.} In face restoration, reference images \cite{Tao2024face, ding2024restoration, li2022learning} provide essential priors for recovering detailed and high-frequency facial features. These priors guide the model to generate realistic restorations. Reference-based methods can be divided into two types: single reference and gallery-based images. The single reference approach \cite{Tao2024face} uses one image to guide the restoration, focusing on enhancing facial details from a specific reference. In contrast, the gallery-based approach \cite{ding2024restoration} leverages multiple reference images, offering more diverse information and stronger constraints, improving the accuracy and realism of the restoration. Despite the effectiveness \cite{ding2024restoration} using gallery photos of users, it still requires multi-stage training and fine-tuning on each user during inference.

\begin{figure*}[t]
\centering
\includegraphics[width=2.0\columnwidth]{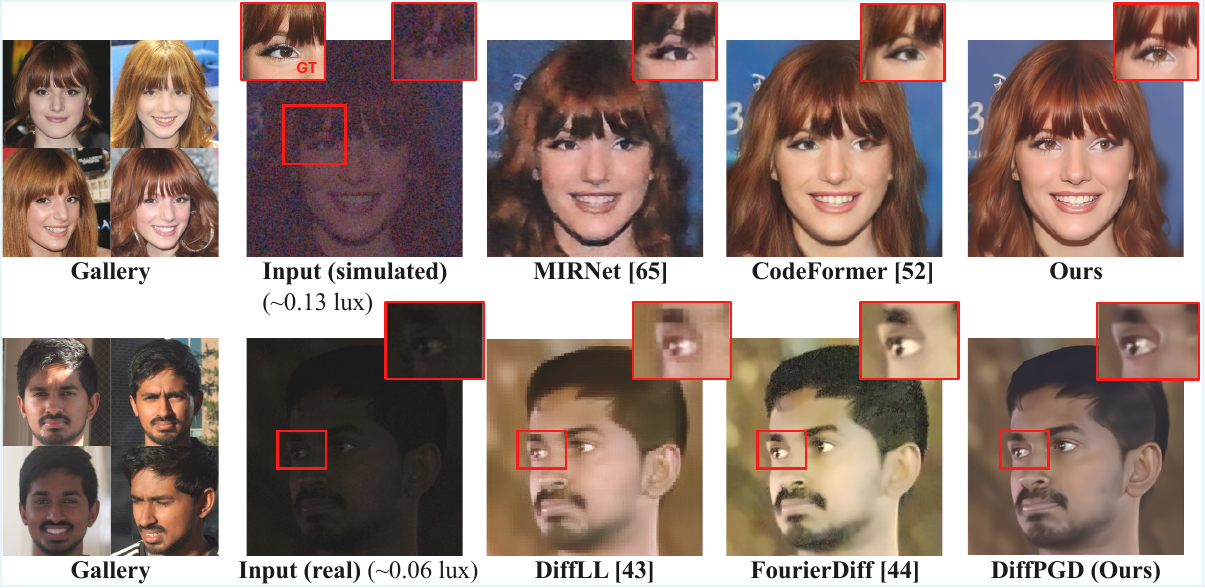}
\setlength{\belowcaptionskip}{-10pt}
\caption{Preview of the visual results. Using gallery photos from a user's smartphone, we can restore low-light, noisy facial images. Our method produces finer details and better identity compared to state-of-the-art low-light and face restoration approaches like MIRNet \cite{zamir2020learning}, DiffLL \cite{jiang2023low}, GFP-GAN \cite{wang2021towards}, FourierDiff \cite{Lv2024FourierPD}, and CodeFormer \cite{zhou2022towards}. Please zoom in for the best visual experience.}
\label{fig:first_figure}
\end{figure*}

The summary comparison between our method and previous methods is summarized in Table  \ref{tab:compact_capability}, and the preview of the visual results can be found in Fig. \ref{fig:first_figure}.


\section{Method}
\label{sec:method}
Our Proposed method is a generative approach based on diffusion models (Section \ref{subsec:diffusion}). However, a naive diffusion model does not support a personalized gallery as prior. The core innovation of our proposed method is the introduction of a physical buffer that can be extracted from the personalized gallery (Section \ref{subsec:why_gallery} and Section \ref{subsec:extract_pb}), and then integrated into the diffusion models to achieve the target denoising (Section \ref{subsec:model_framework}).

\subsection{Denoising Diffusion Model}
\label{subsec:diffusion}

Diffusion models~\cite{SohlDickstein2015DeepUL, ho2020ddpm} approximate a complex data distribution \(q(\mathbf{x}_0)\) by learning to reverse a gradual noising process that transforms the data into a Gaussian distribution \(\mathcal{N}(0, \mathbf{I})\). Through this iterative denoising process, they have become the state of the art in generative modeling, outperforming GANs and VAEs in stability and visual fidelity.

\noindent\textbf{Forward diffusion.}
Given a clean image \(\mathbf{x}_0 \sim q(\mathbf{x}_0)\), the forward process progressively corrupts it with Gaussian noise of increasing variance \(\{\beta_t\}_{t=1}^T\), forming a Markov chain:
\begin{equation}
q(\mathbf{x}_{t} \mid \mathbf{x}_{t-1}) =
\mathcal{N}\!\big(\mathbf{x}_t; \sqrt{1 - \beta_t}\mathbf{x}_{t-1},\, \beta_t \mathbf{I}\big).
\end{equation}
After \(T\) steps, the image becomes nearly indistinguishable from Gaussian noise.  
Using properties of Gaussian composition, the sample at an arbitrary step can be expressed as
\begin{equation}
\mathbf{x}_t = \sqrt{\bar{\alpha}_t}\mathbf{x}_0 + \sqrt{1 - \bar{\alpha}_t}\,\epsilon,
\quad \epsilon \sim \mathcal{N}(0, \mathbf{I}),
\end{equation}
where \(\bar{\alpha}_t = \prod_{s=1}^{t}(1 - \beta_s)\) controls the remaining signal-to-noise ratio.  
When \(\bar{\alpha}_t\) approaches zero, the original signal is nearly destroyed.  
Typical choices of the variance schedule include the linear schedule of Ho \textit{et al.}~\cite{ho2020ddpm, dhariwal2021diffusion} and the cosine schedule of Nichol \textit{et al.}~\cite{Nichol2021ImprovedDD}, which better distribute the noise levels across timesteps and improve sample quality.

\newcommand{\cmark}{\checkmark}
\newcommand{\xmark}{\ding{55}}
\newcommand{\na}{---}

\begin{table}[t]
\scriptsize
\caption{Compared to previous related methods, our approach performs joint low-light enhancement and denoising \textbf{(low-light)}, effectively recovering identity-consistent faces even under severe degradation \textbf{(ID-consistent)}. By leveraging the user's gallery, our method supports personalized restoration \textbf{(Personalized)},  and is scalable to multiple users without requiring fine-tuning when switching users \textbf{(No fine-tune)}.}
\centering
\setlength{\tabcolsep}{5pt}
\renewcommand{\arraystretch}{1.08}
\resizebox{\linewidth}{!}{%
\begin{tabular}{l|c|c|c|c}
\toprule
\textbf{Method} & \textbf{low-light} & \textbf{ID-consistent} & \textbf{Personalized} & \textbf{No fine-tune} \\
\midrule
DiffLL \cite{jiang2023low}        & \xmark & \xmark & \na    & \na \\
FourierDiff \cite{Lv2024FourierPD}  & \xmark & \xmark & \na    & \na \\
MIRNet \cite{zamir2020learning}        & \cmark & \xmark & \na    & \na \\
SNR-aware \cite{xu2022snr}        & \xmark & \xmark & \na    & \na \\
GFPGAN \cite{wang2021towards}        & \cmark & \xmark & \xmark & \cmark \\
CodeFormer \cite{zhou2022towards}    & \cmark & \xmark & \xmark & \cmark \\
Ding \textit{et al.} \cite{ding2024restoration}    & \cmark & \xmark & \cmark & \xmark \\
PFStorer \cite{Varanka2024gallery}      & \cmark & \xmark & \cmark & \cmark \\
\rowcolor{gray!20} \textbf{DiffPGD (Ours)} 
              & \cmark & \cmark & \cmark & \cmark \\
\specialrule{\heavyrulewidth}{0pt}{0pt}
\end{tabular}
}
\label{tab:compact_capability}
\end{table}

\noindent\textbf{Reverse denoising.}
The generative (reverse) process learns to invert the corruption by gradually removing noise from \(\mathbf{x}_T \sim \mathcal{N}(0, \mathbf{I})\) to recover \(\mathbf{x}_0\).  
A neural network \(\epsilon_{\theta}\) is trained to predict the noise component added at each step, thereby parameterizing the reverse conditional
\begin{equation}
p_{\theta}(\mathbf{x}_{t-1} \mid \mathbf{x}_t) =
\mathcal{N}\!\big(\mathbf{x}_{t-1}; \mu_{\theta}(\mathbf{x}_t, t, \mathbf{c}), \sigma_t^2 \mathbf{I}\big),
\end{equation}
where \(\mathbf{c}\) denotes any optional conditioning signal.  
The network is optimized via the simplified denoising objective~\cite{ho2020ddpm, rombach2022high} derived from the variational lower bound:
\begin{equation}
\label{eq:loss_diff_simple}
\mathcal{L} =
\mathbb{E}_{\mathbf{x}_0,\, \mathbf{c},\, \epsilon,\, t}
\big[\|\epsilon - \epsilon_\theta(\mathbf{x}_t, t, \mathbf{c})\|_2^2\big],
\end{equation}
where \(t\) is uniformly sampled from \(\{1, \ldots, T\}\).  
Intuitively, \(\epsilon_\theta\) learns a vector field that points from noisy samples back to the data manifold at different noise scales.

\noindent\textbf{Sampling process.}
At inference, given a noisy image \(\mathbf{x}_t\), the model iteratively refines it by sampling from the learned reverse process:
\begin{equation}
\mathbf{x}_{t-1} =
\frac{1}{\sqrt{1 - \beta_t}}
\left(
\mathbf{x}_t - \frac{\beta_t}{\sqrt{1 - \bar{\alpha}_t}}
\epsilon_\theta(\mathbf{x}_t, t, \mathbf{c})
\right) + \sigma_t z,
\end{equation}
where \(z \sim \mathcal{N}(0, \mathbf{I})\) and typically \(\sigma_t^2 = \beta_t\).  
This iterative denoising gradually reconstructs the clean data \(\mathbf{x}_0\) from pure noise \(\mathbf{x}_T\).

This forward–reverse framework provides a simple yet powerful formulation that unifies generative modeling, denoising, and Bayesian inference. By adjusting the conditioning signal \(\mathbf{c}\), the same architecture can handle diverse tasks such as image restoration, super-resolution, and text-guided generation. In restoration settings, \(\mathbf{c}\) often represents the degraded observation or auxiliary features (e.g., edge, illumination, or identity), and the reverse process recovers a clean image consistent with this condition.

\subsection{Why Combining Gallery and Physical Buffer?}
\label{subsec:why_gallery}

\textbf{Motivation.}
After introducing the diffusion framework in Section~\ref{subsec:diffusion}, we recall that a conditional diffusion model requires additional information \(\mathbf{c}\) to constrain the reverse process.
For personalized restoration, such a condition must encode the subject’s identity to prevent hallucination.
We therefore investigate two possible priors for identity preservation:
(1) the \emph{gallery prior}, and (2) the \emph{physical-buffer prior}.

A \emph{gallery} refers to a small set (typically 3--6) of high-quality face photos of the target person, captured under different poses and lighting conditions.
These images are readily available from user photo collections.
They contain rich identity cues, but are also affected by background, expression, and illumination variations.

A \emph{physical buffer} refers to an intermediate facial representation---in our case the \emph{albedo} and \emph{surface normal}---that describes the physical properties of a person’s face independent of lighting and shading. \emph{Albedo} captures the intrinsic color of an object's surface—that is, the subject's color properties without any lighting effects. It reveals a person's facial appearance, e.g. eyebrows, lip coloration, and other distinct texture elements can reflect individual identity traits. \emph{Surface normal} encodes the surface orientation at each pixel, which provides hints about the underlying shape characteristics. These buffers can be extracted using pretrained face reconstruction networks \cite{feng2021learning, Zhang2021lap}.
Compared to the gallery, they provide disentangled, lighting-independent identity features, but are less directly visual.

\begin{figure}[t]
\centering
\includegraphics[width=1.0\columnwidth]{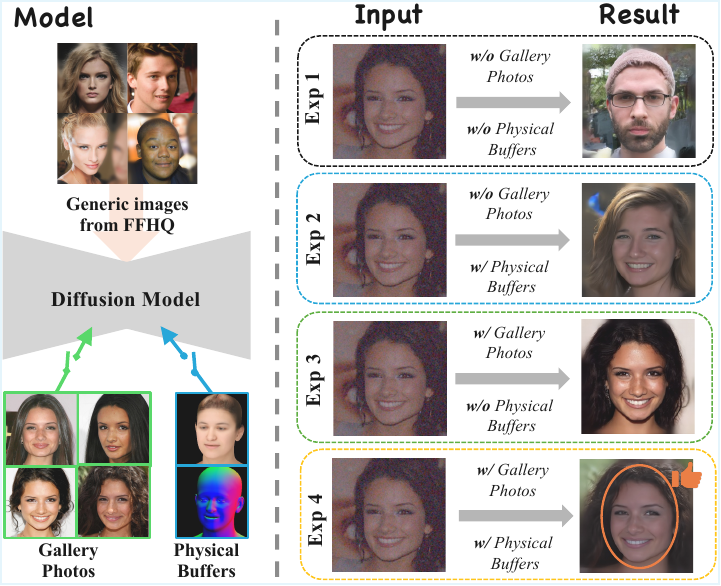}
\caption{Incorporating gallery photos and physical buffers enhances identity preservation. \textbf{Exp 1}: Trained on generic images only~$\rightarrow$~outputs a generic face. \textbf{Exp 2}: Adds physical buffer conditioning~$\rightarrow$~captures some user-specific features. \textbf{Exp 3}: Fine-tuned with gallery photos~$\rightarrow$~generates an arbitrary user face.  \textbf{Exp 4}: Combines gallery fine-tuning and physical buffer~$\rightarrow$~restores the expected user face.}
\label{fig:path_idea}
\end{figure}

\textbf{Controlled Comparison.}
To analyze the complementary effects of these priors, we conduct four controlled generative experiments (Fig.~\ref{fig:path_idea}). 


\begin{itemize}
    \item \textbf{Experiment 1}: Model trained solely on generic images from the FFHQ dataset \cite{Karras2018ASG}.
    \item \textbf{Experiment 2}: Model trained on generic images, while conditioned on the facial physical buffer extracted from the noisy input image.
    \item \textbf{Experiment 3}: Model trained on generic images, then fine-tuned with the user’s gallery photos.
    \item \textbf{Experiment 4}: Model trained on generic images, fine-tuned with the user’s gallery photos, and conditioned on the facial physical buffer extracted from the input image.
\end{itemize}

We tested these trained models on a mildly low-light noisy photo of a user, and the results are presented in the Fig.~\ref{fig:path_idea} (right). The model from Exp~1 produces random faces, as expected from a purely generative purpose. In contrast, the model from Exp~2 uses some coarse face physical buffers extracted from the input image \cite{feng2021learning} as condition. The result from Exp~2 yields faces with user-like features (e.g., nose shape, mouth shape and skin tone). Hence, even just some coarse physical buffers extracted from a degraded photo can convey some key identity cues. Exp~3 outputs distinct faces of the same user, compared to Exp~1, showing that fine-tuning with gallery photos transfers identity information. Finally, Exp 4 yields the best results, showcasing that the \textit{collaborative} use of gallery photos and facial physical prior can achieve the optimal outcomes.

\subsection{Extracting Physical Buffer from Photo Gallery}
\label{subsec:extract_pb}

\begin{figure}[h]
\centering
\includegraphics[width=1.0\linewidth]{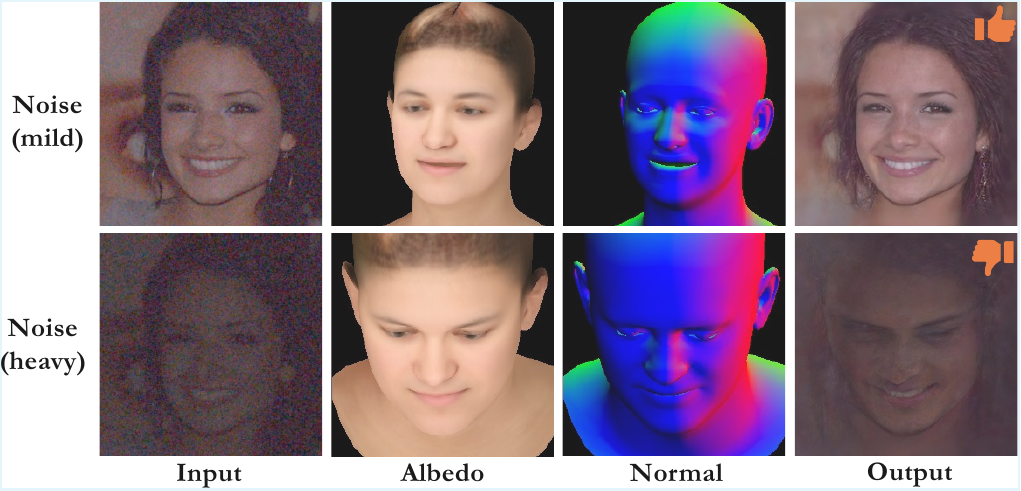}
\caption{For input images affected by mild low-light noise,  physical buffers can be accurately extracted from input. However, when encountering the input images degraded by severe low light and noise, precise physical buffers are difficult to obtain from input.}
\label{fig:deca}
\end{figure}

While Fig. \ref{fig:path_idea} shows us a design direction, there are still two fundamental limitations we need to overcome:
\begin{enumerate}
    \item[(i)] \textbf{Fine-tuning:} Extracting identity via fine-tuning requires significant time and must be repeated for each user \cite{ding2024restoration}. Additionally, the small size of a gallery compared with a large pre-trained model can lead to overfitting.
    \item[(ii)] \textbf{Unreliable buffers from degraded inputs:} As studied in the face reconstruction literature, low-quality inputs consistently present additional challenges \cite{feng2021learning}. Fig.~\ref{fig:deca} shows that extracting albedo and normal maps directly from degraded inputs becomes unreliable as the noise level increases, often leading to incorrect physical buffers, resulting in poor reconstruction. Moreover, features captured from a single image are coarse and limited, not to mention from a degraded one.
\end{enumerate}


\textbf{Our approach}: Extract Physical Buffer from the gallery. Based on the preceding analysis, we propose extracting physical buffers directly from the clean gallery photos instead of noisy input image. This approach offers significant benefits: The gallery provides diverse but unstructured identity evidence, while the physical buffer offers structured, lighting-independent representations.
By deriving the physical buffers from the clean gallery, we combine these complementary strengths—retaining the gallery’s diversity while achieving physically consistent and accurate facial identity features of the target person.
The resulting ID-consistent buffers remain reliable even under severe noise and serve as a strong condition for the diffusion-based restoration process.



By conditioning on the identity physical buffers directly from existing clean gallery photos, we eliminate the need to fine-tune the model with gallery images and avoid the issue of incorrect buffer extraction. Consequently, the model is better equipped to preserve the target person's identity and facial features, even under severe degradation.

\begin{figure}[t]
\centering
\includegraphics[width=1.0\columnwidth]{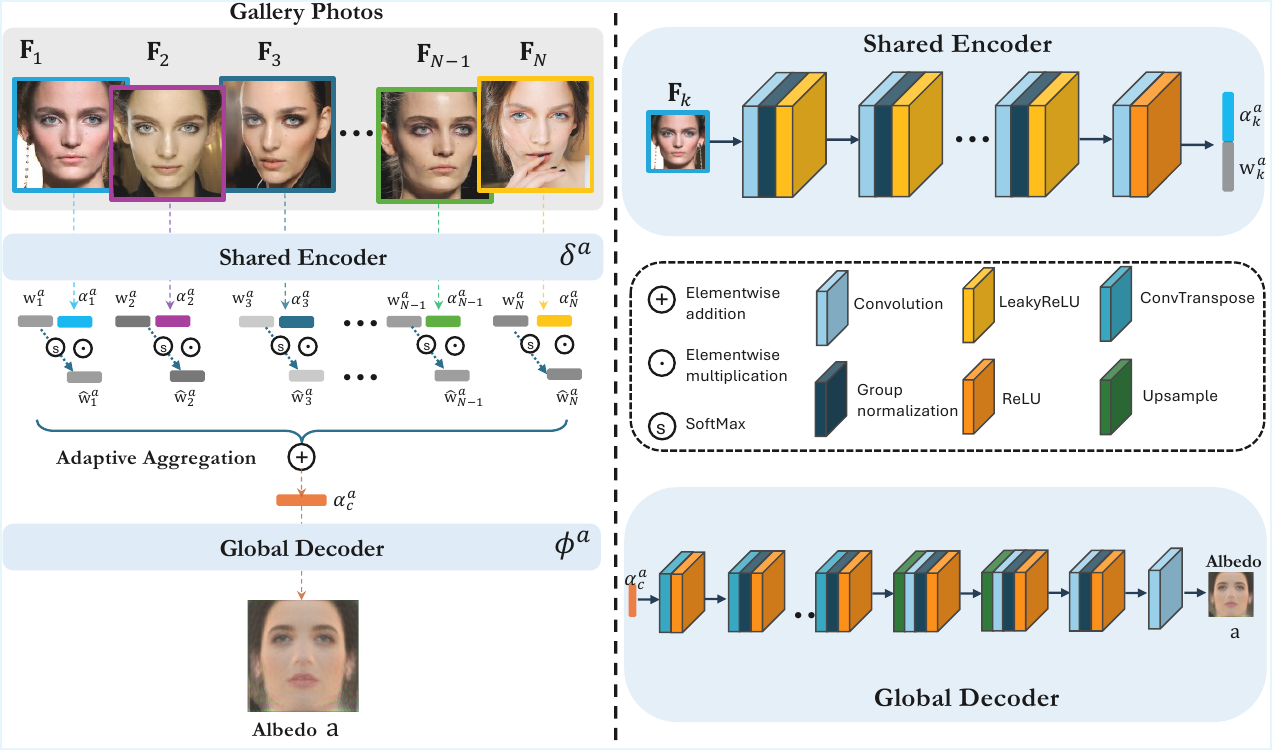}
\caption{ID-consistent physical buffer extractor. The extractor gets the ID physical buffers extracted from target person's gallery photos.  It contains a shared encoder across multiple images and a global decoder for predicting the consistent representation. Note we only show albedo here, and the normal extractor follows the similar strategy. The module weights are initialized from the aggregation network in \cite{Zhang2021lap} and frozen during training.}
\label{fig:pb_extractor}
\end{figure}

\textbf{How to extract physical buffer from gallery?} However, extracting physical buffers from a set of gallery photos is more challenging than extracting from a single input. The complication arises because gallery photos of the same person are often taken under varying conditions, such as different scenes and poses. To ensure these buffers comprehensively represent the person’s identity and facial properties, they must be aligned to maintain identity consistency. Inspired by LAP \cite{Zhang2021lap}, we used an aggregation network as the physical buffer extractor module (shown in Fig. \ref{fig:pb_extractor}, to extract the \emph{ID-consistent} physical buffers, which adaptively combines facial features from multiple images to learn consistent geometry and texture representations, and generates ID-consistent albedo and normal maps from a photo collection of the same individual.  

\textbf{Details of the proposed scheme}. As illustrated in Fig. \ref{fig:pb_extractor}, the physical buffer extractor gets the ID physical buffers  (albedo \(\mathbf{a}^i\) and normal \(\mathbf{n}^i\)) extracted from person \textit{\textbf{i}}'s gallery photos.  The aggregation network contains a shared encoder \(\delta\) across multiple images and a global decoder \(\phi\) for predicting the consistent representation. To model albedo and normal, two separate aggregation networks, denoted as (\(\delta^a, \phi^a\)) and (\(\delta^n, \phi^n\)), are employed (Fig. \ref{fig:pb_extractor} uses the albedo extraction as example, and the normal extraction follows the similar strategy). 

Given a photo gallery of \textit{N} images \(\{\textbf{F}_k^i\}_{k=1}^N\) of person \textbf{\textit{i}} (we will omit the identity index \textbf{\textit{i}} in the following for simplicity), each image is processed by \(\delta^a\) and \(\delta^n\) to extract its texture and geometry latent codes \(
\mathbf{\alpha}_k^a\), \(\mathbf{\alpha}_k^n\) \(\in\) \([1, 1, c]\), respectively.  Due to the quality of \(\{\textbf{F}_k\}_{k=1}^N\), the importance of each dimension in \(
\mathbf{\alpha}_k^a\), \(\mathbf{\alpha}_k^n\) which reveals how correlated it is to the identity, is supposed to be different. Therefore, to derive a global representation of the identity based on \(\{\mathbf{\alpha}_k^a, \mathbf{\alpha}_k^n\}_{k=1}^N\), an adaptive aggregation method is adopted. More specifically, channel-wise weights \(\mathbf{w}_k^a, \mathbf{w}_k^n\) \(\in\) \([1, 1, c]\) are also learned to represent the importance of each dimension in \(
\mathbf{\alpha}_k^a\), \(\mathbf{\alpha}_k^n\): 
\begin{equation}
    [\mathbf{\alpha}_k^a, \mathbf{w}_k^a] = \delta^a (\textbf{F}_k), \text{ and}\quad [\mathbf{\alpha}_k^n, \mathbf{w}_k^n] = \delta^n (\textbf{F}_k).
\end{equation}

A soft-max function is further used to normalize the weights into \(\{\hat{\mathbf{w}}_k^a\}_{k=1}^N, \{\hat{\mathbf{w}}_k^n\}_{k=1}^N\). Finally, the combined global ID-code \(\mathbf{\alpha}_c^a\) and \(\mathbf{\alpha}_c^n\) for texture and depth are  calculated as:
\begin{equation}
    \mathbf{\alpha}_c^a = \sum_{k=1}^N \mathbf{\hat{w}}_k^a \cdot\mathbf{\alpha}_k^a, \text{ and}\quad \mathbf{\alpha}_c^n = \sum_{k=1}^N \mathbf{\hat{w}}_k^n \cdot\mathbf{\alpha}_k^n.
\end{equation}
Once calculated, the global ID-code \(\mathbf{\alpha}_c^a\) and \(\mathbf{\alpha}_c^n\) are passed to the decoders  \(\phi^a\) and \(\phi^n\), respectively,  to produce the ID-consistent albedo \(\mathbf{a}\) and normal \(\mathbf{n}\):
\begin{equation}
    \mathbf{a} = \phi^a (\mathbf{\alpha}_c^a), \text{ and}\quad  \mathbf{n} = \phi^n (\mathbf{\alpha}_c^n).
\end{equation}
The algorithm for our physical buffers extraction process is shown in Algorithm 1.  

\begin{algorithm}
\caption{Extracting physical buffers from photo gallery}
\begin{algorithmic}[1]
\State \textbf{Input}: A photo gallery of \textit{N} images \(\{\textbf{F}_k^i\}_{k=1}^N\) of person \textbf{\textit{i}} (we omit the identity index \textbf{\textit{i}} in the following for simplicity),
\State $[\mathbf{\alpha}_k^a, \mathbf{w}_k^a] = \delta^a (\textbf{F}_k), \quad [\mathbf{\alpha}_k^n, \mathbf{w}_k^n] = \delta^n (\textbf{F}_k)$
\State $\mathbf{\hat{w}}_k^a = \text{softmax}(\mathbf{w}_k^a), \quad  \mathbf{\hat{w}}_k^n = \text{softmax}(\mathbf{w}_k^n)$
\State $\mathbf{\alpha}_c^a = \sum_{k=1}^N \mathbf{\hat{w}}_k^a \cdot\mathbf{\alpha}_k^a, \quad \mathbf{\alpha}_c^n = \sum_{k=1}^N \mathbf{\hat{w}}_k^n \cdot\mathbf{\alpha}_k^n$
\State $\mathbf{a} = \phi^a (\mathbf{\alpha}_c^a), \quad  \mathbf{n} = \phi^n (\mathbf{\alpha}_c^n)$
\State \Return $\mathbf{a}, \mathbf{n}$
\end{algorithmic}
\end{algorithm}

\begin{figure*}[t]
\centering
\includegraphics[width=1.7\columnwidth]{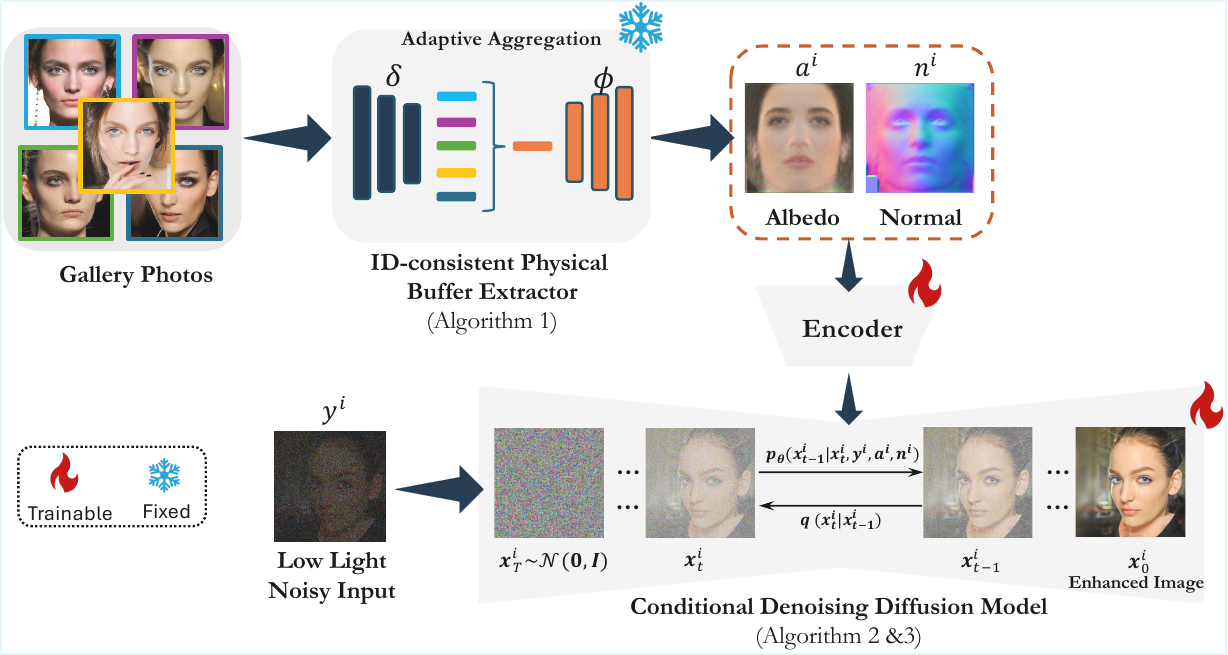}
\setlength{\belowcaptionskip}{-10pt}
\caption{Overview of our method. We leverage ID-consistent physical buffers (albedo and normal) extracted from gallery photos to constrain the generative space during diffusion restoration. The albedo encodes intrinsic skin/appearance while the normal represents facial geometry. In our framework, the output physical buffers isolate the intrinsic ID properties from lighting, shading, and pose, enabling the diffusion model to apply only ID-related information consistently.}
\label{fig:framework}
\end{figure*}


\subsection{Diffusion-based Restoration Model Framework}
\label{subsec:model_framework}

In Section \ref{subsec:extract_pb} we discussed how to extract the physical buffer. Now, we discuss how this physical buffer can be used in the restoration pipeline. We name our final proposed model \textbf{DiffPGD}, which adopts a diffusion-based restoration framework. As illustrated in Fig. \ref{fig:framework}, the model leverages ID physical buffers  (albedo \(\mathbf{a}^i\) and normal \(\mathbf{n}^i\)) extracted from person \textit{\textbf{i}}'s gallery photos as conditions to guide our diffusion model during training to direct the generative process toward a personalized space.  We will drop the superscript \(i\) in the following for simplicity.  

In restoration tasks, such as low-light denoising task solved in this paper, the goal is to recover a high-quality image \(\mathbf{x}_0\) from a low-quality input \(\mathbf{y}\). Unlike a pure generation task, where the output can be any image that follows the distribution of clean images, restoration is constrained by the content of \(\mathbf{y}\). As shown in the conditional denoising diffusion model in Fig. \ref{fig:framework}, the forward/diffusion process (from right to left) gradually adds Gaussian noise, denoted by \(q(\mathbf{x}_t|\mathbf{x}_{t-1})\). Our goal is to reverse the diffusion process (from left to right) by gradually recovering the image from the input Gaussian noise with conditions from different modalities, which corresponds to learning the reverse process of a fixed Markov Chain of length \(T\) conditioned on \(\mathbf{y}\) and the facial physical buffers albedo and normal \((\mathbf{a}, \mathbf{n})\),  extracted from the photo gallery.  When training our model, we adopt a P2-weighted loss format \cite{choi2022perception} with an additional weight term  \(\lambda_t\), and the corresponding objective function measures the discrepancy between the predicted and ground-truth noise, and can be formatted as:
\begin{equation}
\label{eq:loss_diff_restoration}
\mathcal{L} = \lambda_t\mathbb{E}_{\mathbf{x}_0, \mathbf{y}, \mathbf{a}, \mathbf{n}, \epsilon, t}[\|\epsilon - \epsilon_\theta(\mathbf{x}_t, t, \mathbf{y}, \mathbf{a}, \mathbf{n}) \|_2^2] ,  
\end{equation}
According to Eq.~\eqref{eq:loss_diff_restoration}, \(\epsilon_\theta\) takes as input the noisy target image $\mathbf{x}_t$, the time step $t$, and the conditional signal $\mathbf{c}$, which consists of the low-quality input image $\mathbf{y}$ and the extracted physical buffers $(\mathbf{a}, \mathbf{n})$, to estimate the noise \(\epsilon\). \(\lambda_t\) serves as a hyperparameter to adjust the loss weight dynamically across different time step. It assigns higher weights to the loss at levels where the model learns perceptually rich contents while minimal weights to which the model learns imperceptible details.

During inference, the process begins with a pure Gaussian noise image \(\mathbf{x}_T\sim \mathcal{N}(\textbf{\textit{0}}, \textbf{\textit{I}})\), the model learns the conditional transition distribution \(p_\theta(\mathbf{x}_{t-1}|\mathbf{x}_{t}, \mathbf{y}, \mathbf{a}, \mathbf{n})\) and iteratively denoise the image for \(T\) steps, generating the target image \(\mathbf{x}_0\) in the end.  Following the notations defined in Sec. \ref{subsec:diffusion}, the training and inference processes of our model is summarized in Algorithm 2 and 3. 

\begin{algorithm}
\caption{Training our denoising model $\epsilon_\theta$}
\begin{algorithmic}[1]
\State $\mathbf{a, n}$ pre-computed from Algorithm 1
\Repeat
    \State $(\mathbf{x}_0, \mathbf{y}) \sim q(\mathbf{x}_0, \mathbf{y})$
    \State $t \sim \text{Uniform}({1, ..., T})$
    \State $\epsilon \sim \mathcal{N}(0, \mathbf{I})$
    \State $\mathbf{x}_t = \sqrt{\bar{\alpha}_t} \mathbf{x}_0 + \sqrt{1 - \bar{\alpha}_t} \, \epsilon$
    \State Take a gradient descent step on 
    \[
    \nabla_\theta \lambda_t\left\| \epsilon_\theta \left(\mathbf{x}_t, t, \mathbf{y}, \mathbf{a}, \mathbf{n} \right) - \epsilon \right\|_2^2
    \]
\Until{converged}
\end{algorithmic}
\end{algorithm}

\begin{algorithm}
\caption{Inference in $T$ iterative refinement steps}
\begin{algorithmic}[1]
\State $\mathbf{x}_T \sim \mathcal{N}(0, \mathbf{I})$
\State $\mathbf{a, n}$ pre-computed from Algorithm 1
\For{$t = T, \ldots, 1$}
    \State $z \sim \mathcal{N}(0, \mathbf{I})$ if $t > 1$, else $z = 0$
    \State $\mathbf{x}_{t-1} = \frac{1}{\sqrt{1 - \beta_t}} \left( \mathbf{x}_t - \frac{\beta_t}{\sqrt{1 - \bar{\alpha}_t}} \epsilon_\theta(\mathbf{x}_t, t, \mathbf{y}, \mathbf{a}, \mathbf{n}) \right) + \sigma_t z$
\EndFor
\State \Return $\mathbf{x}_0$
\end{algorithmic}
\end{algorithm}

Our experiments demonstrate that identity information is effectively preserved with the guidance of these ID-consistent buffers, enabling our model to achieve superior results in human face restoration.


\section{Experiments}
\label{sec:experiments}

\subsection{Dataset} 
\label{subsec:datasets}
\textbf{Training Data.} Since no publicly available datasets exist for low-light noisy face images, we simulate such data using the CelebAMask-HQ dataset~\cite{Lee2020CelebAMaskhq}, the InverseISP algorithm~\cite{li2024efficient}, and the Poisson-Gaussian noise model (\cref{eq:poisson_gaussian}). 

\begin{figure}[t]
\centering
\includegraphics[width=1.0\columnwidth]{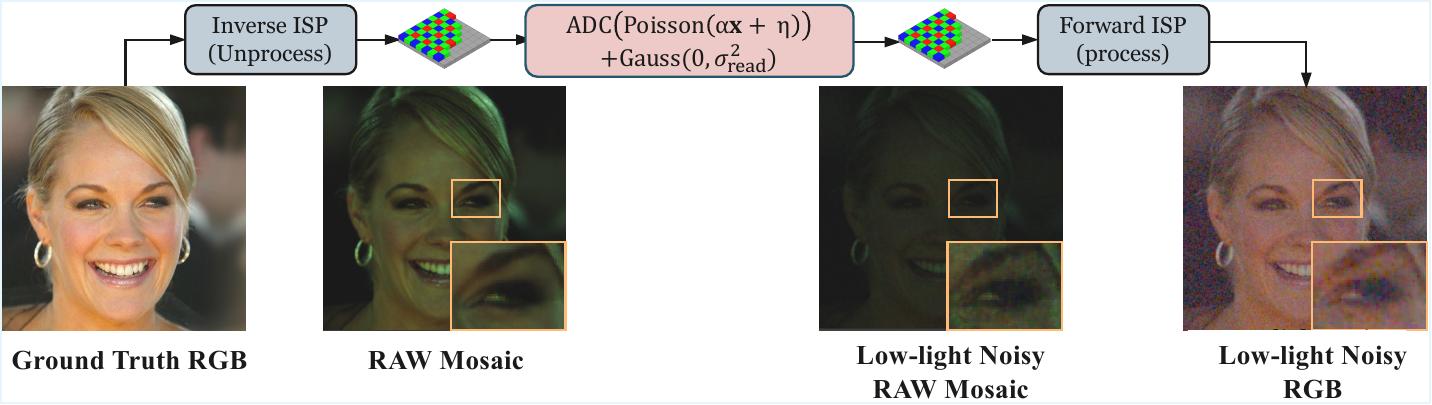}
\caption{Simulation process for generating low-light noisy face images. A ground truth RGB image is first passed through a pre-trained Inverse ISP~\cite{li2024efficient} to obtain a pseudo ground truth RAW mosaic. This is followed by applying a Poisson-Gaussian noise model with varying sensor parameters to simulate a low-light noisy RAW. Finally, a Forward ISP module converts the noisy RAW back to a low-light noisy RGB image.}
\label{fig:simulator}
\end{figure}

As shown in~\cref{fig:simulator}, the first step in the simulation process involves passing the RGB image through a pre-trained Inverse ISP network~\cite{li2024efficient} to generate a pseudo ground truth RAW mosaic image. For visualization purposes, we display the demosaiced version of this image. We utilize the pretrained weights provided~\cite{li2024efficient}, which were obtained by training the network on iPhone camera images for unprocessing the RGB images into RAW format.

The pseudo ground truth RAW image then passed through a Poisson-Gaussian simulator to simulate the low-light noisy RAW mosaic image. During the training process, we vary the camera sensor parameters across a wide range of values to allow the model to learn diverse camera sensor noise profiles. The simulation setup assumes a $12$-bit image sensor with parameters similar to typical DSLR cameras. The parameter ranges are summarized in \Cref{tab:sim_params}.

\setcounter{table}{4}
\begin{table}[h]
    \centering
    \caption{Simulation parameters used in the Poisson-Gaussian Simulator.}
    \label{tab:sim_params}
    \begin{tabular}{lc}
        \Xhline{3\arrayrulewidth}
        \textbf{Parameter} & \textbf{Range} \\ \hline
        Full Well Capacity (FW) & $19000\text{e}^{-} \leq \text{FW} \leq 64000\text{e}^{-}$ \\ 
        Quantum Efficiency ($\mathrm{\eta}$)  & $0.32 \leq \mathrm{\eta} \leq 0.54$ \\ 
        Dark Current ($\mathrm{\sigma_{dark}}$)       & $2.2\text{e}^{-} \leq \mathrm{\sigma_{dark}} \leq 11.7\text{e}^{-}$ \\ 
        Read Noise ($\mathrm{\sigma_{read}}$)        & $2.2\text{e}^{-} \leq \mathrm{\sigma_{read}} \leq 10.8\text{e}^{-}$ \\ 
        photo level ($\mathrm{\alpha}$) & $13 \,\text{ppp} \leq \mathrm{\alpha} \leq 65 \text{ppp}$ \\ \Xhline{3\arrayrulewidth}
    \end{tabular}
\end{table}


\textbf{Testing Data.} To test DiffPGD, we utilize both simulated and real-world captured low-light noisy data. 

For the simulated testing data, similar to the training data simulation, the simulation setup assumes a $12$-bit image sensor with parameters similar to that of a DSLR camera: a full well capacity (FW) of $36000\text{e}^{-}$, quantum efficiency ($\mathrm{\eta}$) of $0.42$, dark current of $\mathrm{\sigma_{dark}}=7\text{e}^{-}$, and read noise of $\mathrm{\sigma_{read}}=6\text{e}^{-}$. These parameters ensure realistic noise characteristics for evaluating model performance. 

We also captured 100 RAW-format face images as real-world test dataset using a Sony $\alpha6400$ DSLR camera paired with a Sony 18-135mm lens at night. These images feature diverse low-light and dark backgrounds, various angles and poses, as well as different focal lengths, apertures, ISO values, and shutter speeds to reflect the rich characteristics of real-world low-light photography. We cropped these images into 256×256 face patches.

\subsection{Implementation Details}
Our denoising model is built upon the ADM architecture~\cite{dhariwal2021diffusion}. As illustrated in Fig.~\ref{fig:framework}, we modify the original architecture so that the model can take concatenated degraded images as conditional inputs. In addition, we employ an encoder (ResNet-18~\cite{he2016deep}) to extract a global latent code from the target individual’s physical buffers, which are obtained from the gallery photos. Following the FiLM-style \cite{Perez2017FiLMVR} modulation used for time embeddings in ADM~\cite{dhariwal2021diffusion}, this latent code modulates the UNet features in each residual block via scale–and–shift or additive transforms, thereby injecting identity information into the diffusion model. The model parameters are configured as follows: the number of channels is set to 128, channel multiple is (1, 1, 2, 2, 4, 4), head channels are 128, attention resolution is 16, dropout rate is 0.1, diffusion steps are 1000. For additional architectural details, refer to \cite{dhariwal2021diffusion}.

The ID-consistent physical extractor is directly initialized from the pre-trained weights of the aggregation network in LAP \cite{Zhang2021lap}, without the need of retraining.

For training our model, we use the loss function defined in Eq.~\eqref{eq:loss_diff_restoration}. The image size is 256 \(\times\) 256. During the training stage, we train our model on the CelebAMask-HQ dataset \cite{Lee2020CelebAMaskhq}, comprising 30,000 images distributed across 4,516 unique identities (IDs). We use 3318 IDs for training and 200 IDs for simulated testing. Within each ID, there are 3–20 images. The number of gallery images we used per person varies from 3 to 6, with most of them around 5. During inference, we used 3 to 6 images per person for the simulated test set, and 6 gallery images per person for the real-captured test set.

The model is trained using the Adam optimizer \cite{kingma2014adam} with a learning rate of \text{$10^{-4}$}. The weight decay is set to 0.  Training is performed for 200,000 iterations with a batch size of 32, resulting in a total of 6,400,000 samples seen by the model. During inference, we adopt the DDIM method \cite{song2020denoising} with 200 denoising steps. All experiments are performed on a single NVIDIA A100 GPU.

\subsection{Performance in Real-Captured Scenarios}
Testing model performance in real-life low-light scenarios is crucial and challenging, as real-world conditions often involve uncontrollable factors compared to simulated ones. Since no existing data set addresses the restoration of real-life low-light noisy face photos, we captured photos of real persons under indoor and outdoor low-light scenarios, our procedure is described in Section \ref{subsec:datasets}.  

We compare our method with several state-of-the-art methods: UTV-Net~\cite{zheng2021adaptive}, GFPGAN\cite{wang2021towards}, SNR-aware\cite{xu2022snr}, MIRNet\cite{zamir2020learning},  DiffLL\cite{jiang2023low}, Codeformer \cite{zhou2022towards}. These works aim to restore high-quality images from degraded data that has been affected by low light, noise, or both, which are close to the subject of our research. Particularly, GFPGAN\cite{wang2021towards} and Codeformer\cite{zhou2022towards} are designed specifically for face image restoration, which is highly competitive with our method. We re-trained/fine-tuned all the baselines with the same training dataset as ours for fair comparison.

\begin{figure*}[t]
\centering
\includegraphics[width=2.0\columnwidth]{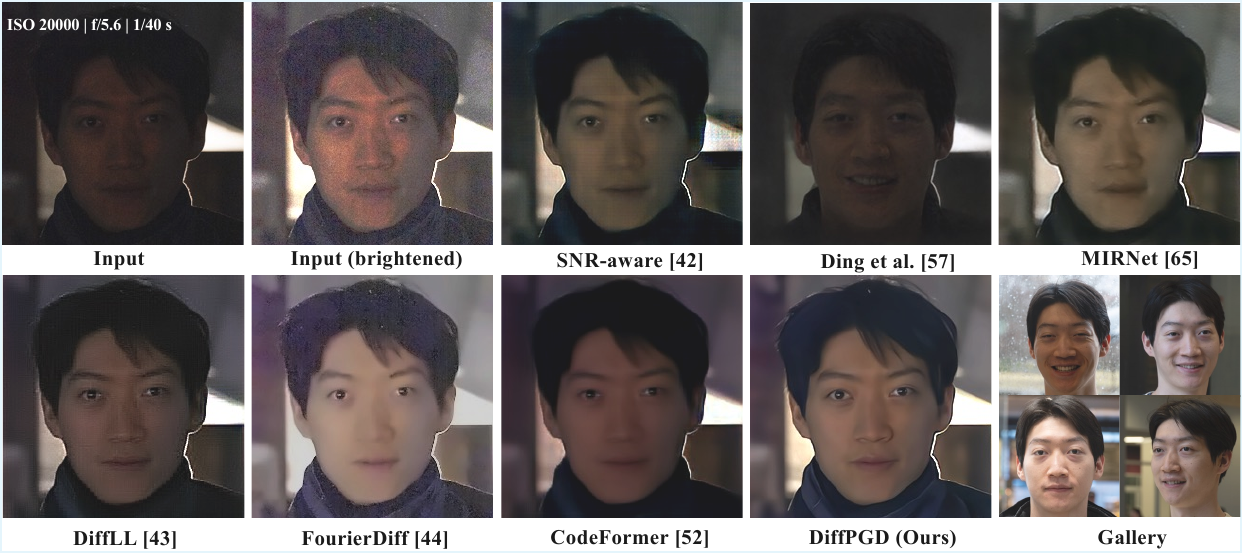}
\caption{Visual comparison on real cases. The first image presents the input images along with their capture parameters. In the second image, we enhance the brightness to better visualize the noise present in the input. Please zoom in for the best visual experience.}
\label{fig:real_test_1}
\end{figure*}

\begin{figure*}[t]
\centering
\includegraphics[width=2.0\columnwidth]{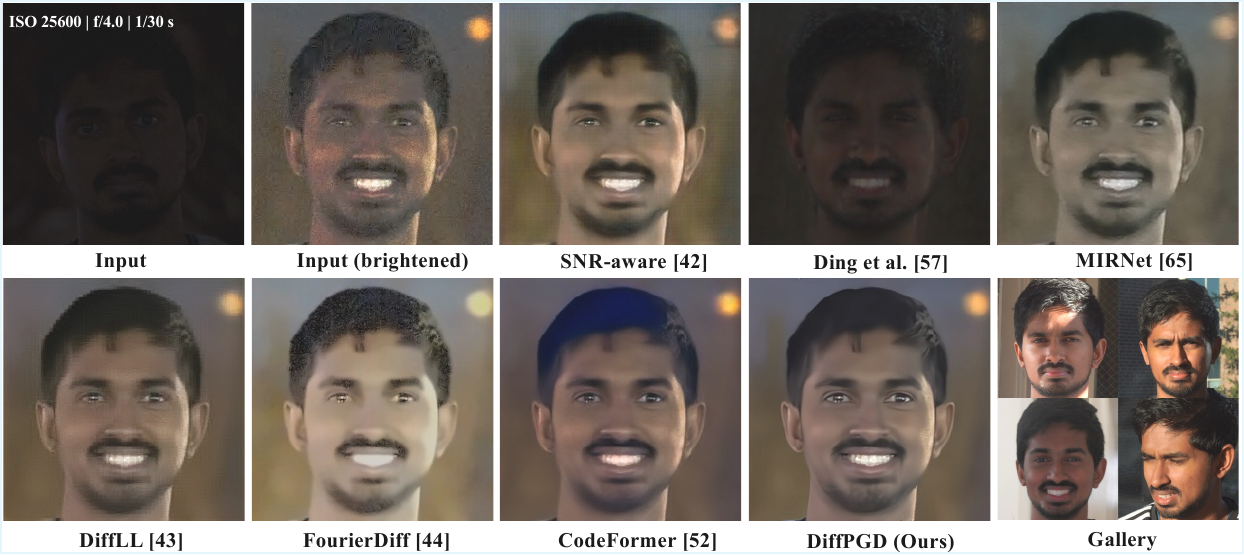}
\caption{Visual comparison on real cases. The first image presents the input images along with their capture parameters. In the second image, we enhance the brightness to better reveal the noise in the input. Please zoom in for the best visual experience.}
\label{fig:real_test_2}
\end{figure*}

\begin{figure*}[t]
\centering
\includegraphics[width=2.0\columnwidth]{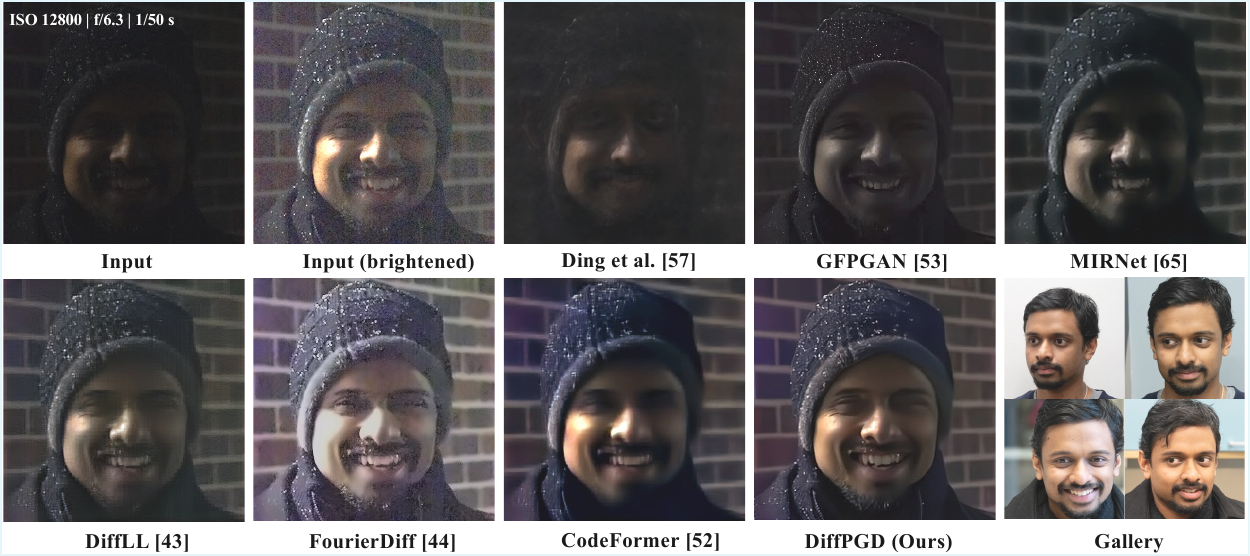}
\caption{Visual comparison on real cases. The first image presents the input images along with their capture parameters. In the second image, we enhance the brightness to better reveal the noise inside the input. Please zoom in for the best visual experience.}
\label{fig:real_test_3}
\end{figure*}

Quantitative and qualitative results are presented. To assess the quality of restored images, we employ the Fréchet Inception Distance (FID) \cite{Heusel2017FID} and the Kernel Inception Distance (KID) \cite{Binkowski2018KID}, as these metrics offer a robust evaluation aligned with human perception. Since ground-truth images are unavailable for real-captured cases, we compute FID and KID scores between the model's output and the corresponding subjects' gallery photos. The results are shown in Table \ref{tab:real_compare}. Our method achieves the best FID score and the second-best KID score. When examining the visual results presented in Figs. \ref{fig:real_test_1}, \ref{fig:real_test_2} and \ref{fig:real_test_3}, our method produces images with the highest perceptual quality and level of detail. The restored images exhibit more natural skin tones and well-preserved facial identity features, such as eye shape, eyebrows, and lips.

\begin{figure*}[t]
\centering
\includegraphics[width=2.0\columnwidth]{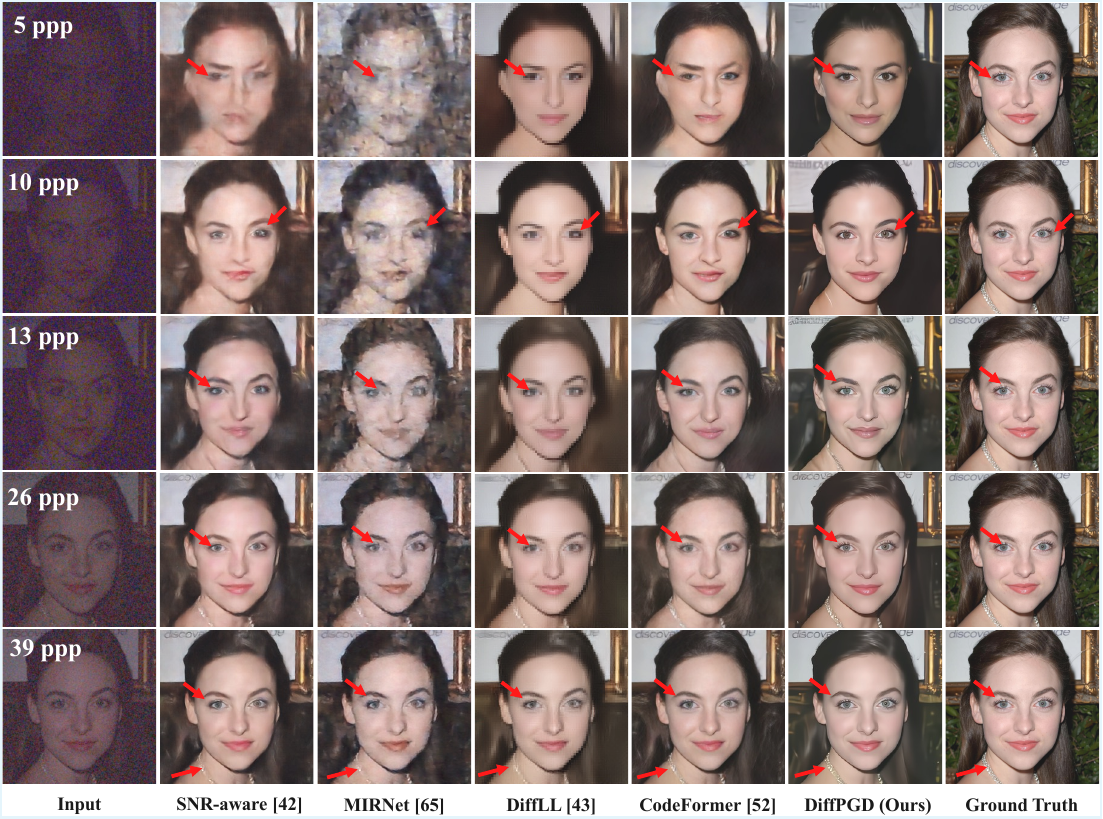}
\caption{Visual comparison on the simulated cases. The lower the ppp, the more severe the degradation is. Our method maintains good identity consistency even at extremely low ppp and exhibits fewer artifacts. Please zoom in for the best visual experience.}
\label{fig:ppp_trend}
\end{figure*}

Since our work focuses on face images, we also check on the identity preservation performance. Besides the image quality metrics, we adopt the identity score, which computes the cosine similarity of face embeddings extracted by the FaceNet model \cite{Schroff2015FaceNetAU} pretrained on the VGGFace2 dataset \cite{Cao2017VGGFace2AD}, to evaluate identity consistency. Again, since ground truth images are unavailable for real-captured cases, we calculate the identity (ID) score by averaging the ID scores between the model's output images and each gallery photo of the corresponding subjects.  The results are shown in Table \ref{tab:real_compare}. Our method has the highest identity score, and the visual results are consistent with the quantitative metric results, maintaining the highest consistency in facial identity. The results demonstrate the ability of our proposed DiffPGD model on real low light denoising and enhancement tasks. 

\begin{table}[ht]
    \centering
    \caption{Comparisons with state-of-the-art works on the real-captured test cases.}
    \resizebox{0.45\textwidth}{!}{
    \small
    \begin{tabular}{c|c|c|c}
        \Xhline{4\arrayrulewidth}
        Methods & FID $\downarrow$ & KID $\downarrow$ & ID Score $\uparrow$ \\
        \hline
        SNR-aware\cite{xu2022snr} & 358.03 & 0.3206 & 0.56 \\
        
        UTVNet\cite{zheng2021adaptive} & 360.13 & 0.2304 & 0.42 \\
        
        MIRNet\cite{zamir2020learning} & 274.57 & \textbf{0.1732} & 0.63 \\

        DiffLL\cite{jiang2023low} & 301.13 & 0.2496 & 0.69 \\

        GFPGAN\cite{wang2021towards} & 276.30 & 0.2023 & 0.30 \\

        CodeFormer\cite{zhou2022towards} & 295.68 & 0.2158 & 0.61 \\

        Ding \textit{et al.} \cite{ding2024restoration} & 295.85 & 0.2438 & 0.24 \\

        FourierDiff\cite{Lv2024FourierPD} & 303.78 & 0.2383 & 0.70 \\
        
        \rowcolor{Gray}\textbf{Ours} & \textbf{270.14} & 0.1765  & \textbf{0.72} \\
        \Xhline{4\arrayrulewidth}
    \end{tabular}}
    \label{tab:real_compare}
\end{table}

\subsection{Performance in Simulated Scenarios}
In simulated scenarios, we use the 200 test images which correspond to the 200 test IDs from CelebAMask-HQ and send them into our simulator to generate the test set. We simulated various low-light noisy levels by setting different photons per pixel (ppp), i.e., the average number of photons arriving at the sensor within the exposure time. A low photon level increases the impact of read noise and dark current on the final output, thereby decreasing the quality of the reconstructed images.


Quantitative results are provided in Table \ref{tab:simulate_compare}. The image restored by our method has the highest quality and more details, meanwhile preserving the facial identity and maintaining the ID consistency. From the table, we could see that CodeFormer surpasses us on PSNR value. Fig. \ref{fig:ppp_trend} illustrates a visual comparison between our method and state-of-the-art works under inputs ranging from 5 to 39 ppp, which corresponds to $\sim$0.049 lux to 0.384 lux, estimated at the sensor side assuming typical low-light CMOS settings with pixel pitch in the range of 2–3 $\mu\text{m}$, quantum efficiency 0.3–0.54, and short integration time on the order of 10–30 ms. We observe that CodeFormer tends to generate smoother images while we could preserve better high-frequency details. Furthermore, our method demonstrates superior identity preservation even at low resolutions, such as 5 ppp to 13 ppp ($\sim$0.049 lux to 0.128 lux), while exhibiting fewer artifacts. The advantage of our model in maintaining identity consistency is particularly evident in heavily degraded cases. The results highlight the strong performance of our proposed DiffPGD model in low-light denoising and enhancement tasks. Like any denoising method, there is a physical limit beyond which our method will break down. This happens when input image is below 5ppp ($\sim$0.049 lux). However, in this extremely low photon condition, we still perform much better than competing methods.

\begin{table*}[ht]
    \caption{Quantitative comparison with state-of-the-art works on CelebAMask-HQ dataset on 5ppp ($\sim$0.049 lux), 10ppp($\sim$0.098 lux), 13ppp($\sim$0.128 lux), 26ppp($\sim$0.256 lux), and 39ppp($\sim$0.384 lux). The lower the ppp, the more severe the degradation is.}
    \centering\resizebox{\textwidth}{!}{
    \Huge
    \begin{tabular}{c|ccc|ccc|ccc|ccc|ccc}
        \Xhline{8\arrayrulewidth}
          \textbf{lux} & \multicolumn{3}{c|}{\textbf{0.049 lux}} & \multicolumn{3}{c|}{\textbf{0.098 lux}} & \multicolumn{3}{c|}{\textbf{0.128 lux}} & \multicolumn{3}{c|}{\textbf{0.256 lux}} & \multicolumn{3}{c}{\textbf{0.384 lux}}\\
        \hline
         Methods
        & ID Score $\uparrow$ & FID $\downarrow$& KID  $\downarrow$
        & ID Score $\uparrow$ & FID $\downarrow$& KID  $\downarrow$
        & ID Score $\uparrow$ & FID $\downarrow$& KID  $\downarrow$
        & ID Score $\uparrow$ & FID $\downarrow$& KID  $\downarrow$
        & ID Score $\uparrow$ & FID $\downarrow$& KID  $\downarrow$\\
        \hline
        SNR-aware\cite{xu2022snr} & 0.3093  & 200.10 &  0.1382 &  0.5580  & 134.44 & 0.0664 & 0.6366 & 120.44 & 0.0526 & 0.7922 & 96.55 & 0.0330  & 0.8547  & 84.36 &  0.0244\\
        
        MIRNet\cite{zamir2020learning} & 0.1256 & 376.98 & 0.4035 & 0.3596 & 320.81 & 0.2719 & 0.4698 & 303.99 & 0.2506 & 0.7068 & 209.01 & 0.1579 & 0.7928 & 161.86 & 0.1106  \\

        FourierDiff\cite{Lv2024FourierPD} & 0.0451 & 384.58 &  0.4545 &  0.2045 & 308.99 & 0.2753 & 0.3233 &  275.99  & 0.2121 & 0.6118 & 202.49 &  0.1462 & 0.7195 & 163.25 & 0.1137  \\
        
        DiffLL\cite{jiang2023low} & 0.3835 & 190.41 &  0.1687 &  0.5496 & 167.77 & 0.1390 & 0.6228 &  155.87  & 0.1244 & 0.7796 & 128.73 &  0.0898 & 0.8388 & 115.59 & 0.0776  \\

        GFPGAN\cite{wang2021towards} & 0.0070 & 388.93 & 0.4256 & 0.1194 & 233.32 & 0.1522 & 0.2851 & 203.81 & 0.1152 & 0.6175 & 123.27 & 0.0624 & 0.7294 & 106.47 & 0.0497 \\

        CodeFormer\cite{zhou2022towards} & 0.3553 & 111.46 & 0.0491 & 0.5867 & 91.13 & 0.0326 & 0.6501 & 84.80 & 0.0281 & 0.7998 &  80.66 &0.0264 & 0.8645 & 64.04 & 0.0167  \\
        
        \rowcolor{Gray}\textbf{Ours} & \textbf{0.4527} & \textbf{85.50} & \textbf{0.0178} & \textbf{0.6236} & \textbf{75.39} & \textbf{0.0135} & \textbf{0.6923} & \textbf{75.65} & \textbf{0.0179} & \textbf{0.8168} & \textbf{68.44} & \textbf{0.0114} & \textbf{0.8719} & \textbf{62.73} & \textbf{0.0112}  \\
        \Xhline{8\arrayrulewidth}
    \end{tabular}}
    \label{tab:simulate_compare}
\end{table*}

\section{Ablation Study}
\label{sec:ablation}
\textbf{ID-Consistent Physical Buffers for Identity Preservation.} In Table \ref{tab:ablation}, we compare the quantitative results of training our model with and without ID physical buffers in real-captured test cases. The results demonstrate that using physical buffers significantly increases the identity score, highlighting the effectiveness of extracting ID physical buffers from a person's existing photo gallery to assist in restoring real-world low-quality photos.

\begin{table}[ht]
    \centering
    \caption{Ablation comparisons of using ID physical buffers on real-captured test cases.}
    \resizebox{0.45\textwidth}{!}{
    \small
    \begin{tabular}{c|c|c|c}
        \Xhline{4\arrayrulewidth}
        Methods & FID $\downarrow$ & KID $\downarrow$ & ID Score $\uparrow$ \\
        \hline
        w/o physical buffers & 281.68 & 0.1814 & 0.68 \\
        
        \rowcolor{Gray}w/ physical buffers & \textbf{270.14} & \textbf{0.1765} & \textbf{0.72} \\
     \Xhline{4\arrayrulewidth}
    \end{tabular}}
    \label{tab:ablation}
\end{table}

\begin{figure}[H]
\centering
\includegraphics[width=0.75\columnwidth]{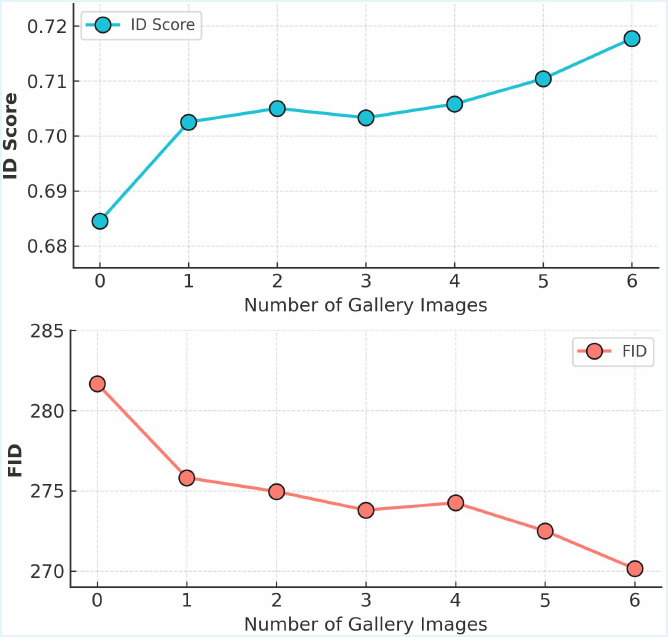}
\caption{The relationship between the identity (ID) score and FID of the restored images and the number of gallery images used to generate the ID physical buffer.}
\label{fig:trend_g}
\end{figure}

\textbf{Effect of the Number of Gallery Images.} The proposed DiffPGD model is conditioned with the ID physical buffers extracted from the user's gallery photos. As stated in the paper, facial features extracted from a single image are inherently limited. In Fig. \ref{fig:trend_g}, we evaluate the effect of the number of gallery images used to generate the ID physical buffer. We analyze it with the real-captured set where 6 gallery images were collected from each subject. The results show that identity preservation improves as the ID physical buffer is derived from a greater number of gallery photos rather than a single image. Additionally, incorporating the ID physical buffer significantly enhances identity preservation compared to not using it.

\begin{figure}[t]
\centering
\includegraphics[width=1\columnwidth]{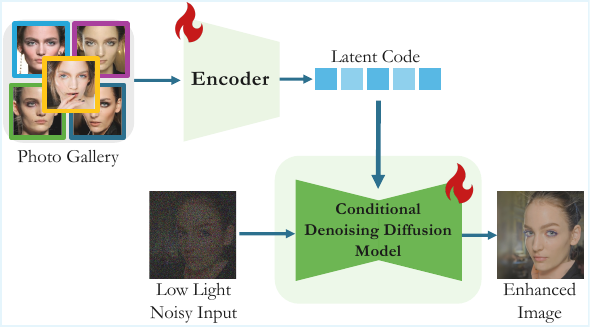}
\caption{A network architecture diagram for identity feature restoration. It extracts latent identity features from the user’s photo gallery directly with an encoder, which are then used to condition a denoising diffusion model.}
\vspace{-3pt}
\label{fig:face_encoder}
\end{figure}

\textbf{Direct-Gallery vs. Physical Buffer?} 
Some previous studies have attempted to extract latent facial feature representations directly from reference photos~\cite{Varanka2024gallery}. To evaluate whether this strategy can effectively capture sufficient facial information to guide the diffusion model, compared with extraction from physical buffers, we conduct an experiment. As illustrated in Fig.~\ref{fig:face_encoder}, instead of deriving features from physical buffers, we directly extract latent codes from the target individual’s gallery photos.

\begin{table}[!h]
    \centering
    \caption{Comparison between ID physical-buffer-based and direct-gallery-based methods on the real-captured photo test set. For a fair comparison, the same gallery set is used for each individual in both the direct-gallery-based and physical-buffer-based approaches.}
    \resizebox{0.48\textwidth}{!}{
    \small
    \begin{tabular}{c|c|c|c|c}
        \Xhline{4\arrayrulewidth}
        Methods & LPIPS $\downarrow$ & SSIM $\uparrow$ & KID $\downarrow$ & ID Score $\uparrow$ \\
        \hline
        w/ direct-gallery & 0.6316 & 0.2574 & 0.0371 & 0.77 \\
        
        \rowcolor{Gray}w/ physical buffers & \textbf{0.5742} & \textbf{0.4175} & \textbf{0.0161} & \textbf{0.83} \\

     \Xhline{4\arrayrulewidth}
    \end{tabular}}
    \label{tab:encoder_PB}
\end{table}

\begin{figure}[t]
\centering
\includegraphics[width=1\columnwidth]{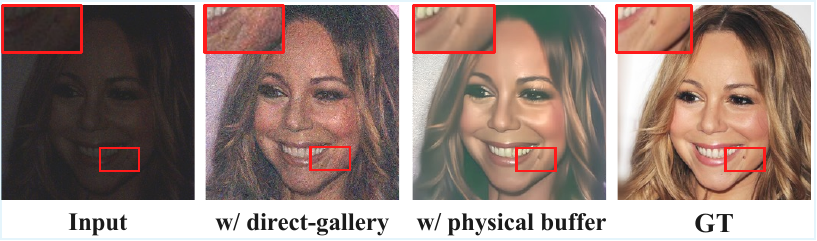}
\caption{Visual comparison on different identity representation strategies.}
\label{fig:PBandFE}
\vspace{-10pt}
\end{figure}

We compare the restored images generated with the direct-gallery-based method with the proposed physical buffer-guided method on our real-captured photo set. This is a special semi-real set, in which we take photos of printed paper photos at night, allowing ground-truth reference for evaluation. The quantitative and visual comparisons are presented in Table \ref{tab:encoder_PB} and Fig. \ref{fig:PBandFE}, respectively. The results demonstrate that guiding the diffusion model using ID physical buffers produces significantly better identity preservation than the direct-gallery-based approach. One major reason for this is that the ID physical buffers serve as robust conditions, enabling the extraction of comprehensive facial features from the gallery. In contrast, the direct-gallery-based method is less focused on facial features and is often influenced by irrelevant factors such as background and hairstyle. Consequently, the latent codes extracted by the encoder are less effective in guiding the diffusion model to the personalized generative space compared to the ID physical buffers.
\vspace{-5pt}
\section{Conclusion}
\label{sec:conclusion}
In this work, we addressed the challenges of low-light facial image restoration by introducing Diffusion-based Personalized Generative Denoising (DiffPGD), a novel approach that leverages user-specific photo galleries to improve denoising performance. By incorporating an identity-consistent physical buffer, we provided a strong prior that enhances the diffusion model's ability to restore degraded images while avoiding the need for model fine-tuning. Our experiments, which included extensive evaluations on multiple challenging datasets, demonstrated that DiffPGD not only outperforms existing diffusion-based denoising methods but also delivers superior quantitative and qualitative results across various testing scenarios. This approach effectively mitigates false content generation in low-SNR conditions while bridging the gap in low-light imaging quality, ultimately paving a way for personalized and more robust image restoration techniques. 

\vspace{5pt}

\begingroup
\setlength{\parskip}{0pt}  
\setlength{\itemsep}{0pt plus 0.3ex}

\bibliographystyle{IEEEtran}
\bibliography{biblio}

\endgroup
\end{document}